# Multiaxis nose-pointing-and-shooting in a biomimetic morphing-wing aircraft

Arion Pons[1,*] and Fehmi Cirak[2]
*Department of Engineering, University of Cambridge, Cambridge CB2 1PZ, UK*

**Modern high-performance combat aircraft exceed conventional flight-envelope limits on maneuverability through the use of thrust vectoring, and so achieve supermaneuverability. With ongoing development of biomimetic unmanned aerial vehicles (UAVs), the potential for supermaneuverability through biomimetic mechanisms becomes apparent. So far, this potential has not been well studied: biomimetic UAVs have not yet been shown to be capable of any of the forms of classical supermaneuverability available to thrust-vectored aircraft. Here we show this capability, by demonstrating how biomimetic morphing-wing UAVs can perform sophisticated multiaxis nose-pointing-and-shooting (NPAS) maneuvers at low morphing complexity. Nonlinear flight-dynamic analysis is used to characterize the extent and stability of the multidimensional space of aircraft trim states that arises from biomimetic morphing. Navigating this trim space provides an effective model-based guidance strategy for generating open-loop NPAS maneuvers in simulation. Our results demonstrate the capability of biomimetic aircraft for air combat-relevant supermaneuverability, and provide strategies for the exploration, characterization, and guidance of further forms of classical and non-classical supermaneuverability in such aircraft.**

### Nomenclature

| | | |
|---|---|---|
| $\boldsymbol{\omega}_i$ | = | body angular velocities, rad/s |
| $\boldsymbol{\omega}$ | = | aircraft reference angular velocity, rad/s |
| $\dot{\mathbf{x}}_i$ | = | body and point velocities, m/s |
| $\mathrm{R}_{j/i}$ | = | rotation matrix from $i$ to $j$ |
| $\boldsymbol{\theta}$ | = | vector of aircraft Euler angles, rad |
| $\boldsymbol{\tau}_n$ | = | vector of wing $n$, $n \in \{$left, right$\}$, Euler angles, rad |
| $\theta$ | = | first Euler angle; pitch, rad |

---

[1,*] PhD student. Present affiliation: JBC Postdoctoral Scholar, Hebrew University of Jerusalem, Giv'at Ram, Jerusalem, Israel. arion.pons@mail.huji.ac.il. Corresponding author.
[2] Reader in Computational Mechanics, fc286@cam.ac.uk

| | | |
|---|---|---|
| $\psi$ | = | second Euler angle; yaw, rad |
| $\gamma$ | = | third Euler angle; roll, rad |
| $\alpha$ | = | aircraft angle of attack, rad |
| $\beta$ | = | aircraft sideslip angle, rad |
| $\Delta\psi$ | = | perturbation in yaw angle, rad |
| $\Omega_i$ | = | transformation matrices between Euler angle rates and angular velocities |
| $m_i$ | = | body masses, kg |
| $I_i$ | = | body rotational inertias, kg m² |
| $\mathbf{L}_i$ | = | body location vectors, m |
| $\upsilon$ | = | set of all system control and structural parameters |
| $\mathcal{W}$ | = | set of wing bodies |
| $\mathcal{F}$ | = | set of fuselage bodies |
| $\mathcal{S}$ | = | set of all bodies |
| $t$ | = | time, s |
| $T$ | = | period, s |
| $K$ | = | kinetic energy, J |
| $\mathbf{L}$ | = | lift force vector, N |
| $\mathbf{D}$ | = | drag force vector, N |
| $\mathbf{M}$ | = | pitching moment vector, Nm |
| $\mathcal{Q}_\mathbf{x}$ | = | generalized force vector, N |
| $\mathcal{Q}_\mathbf{\theta}$ | = | generalized moment vector, Nm |
| $\times$ | = | cross product |
| $[\cdot]_\times$ | = | skew operator |
| $\tan_2^{-1}(\cdot,\cdot)$ | = | 2-argument arctangent function |
| $\text{sgn}(\cdot)$ | = | sign function |
| $\|\cdot\|_2$ | = | Euclidean norm |
| $\propto n$ | = | linearly proportional to $n$ |
| $\mathcal{O}(n)$ | = | order of magnitude of $n$ |
| $E_{n \times n}$ | = | identity matrix of size $n$ |
| $0_{n \times m}$ | = | zero matrix of size $n \times m$ |
| $\rho$ | = | air density, kg/m$^3$ |
| $C_n$ | = | aerodynamic coefficient of force $n$ |
| $\delta_n$ | = | deviation metric in variable $n$ |
| $\phi_{\text{eff},k}$ | = | local station effective angle of attack, rad |
| $\Phi_k$ | = | local station effective angle of attack spectrum, rad |
| $V_k$ | = | local station airspeed, m/s |
| $b_{c,k}$ | = | local station airfoil semichord, m |
| $\omega_k$ | = | local station airfoil spectral frequency, rad/s |
| $\kappa_k$ | = | local station reduced frequency |

**Superscripts**

| | | |
|---|---|---|
| $T$ | = | transpose |
| $\hat{\phantom{x}}$ | = | amplitude |
| $(b),(e),(k)$ | = | resolution in body-fixed, earth, or other defined $(k)$ frames |

**Subscripts**

| | | |
|---|---|---|
| $i$ | = | body index |
| $k$ | = | aerodynamic station index |
| $S$ | = | aircraft reference point $S$ |
| $H$ | = | wing root point $H$ |
| $A/B$ | = | frame $A$ w.r.t. frame $B$ |
| $tr$ | = | trim state |
| $tg$ | = | quasistatic NPAS target |
| $cp$ | = | center-point |
| $p$ | = | perturbation |

## 1. Introduction

Supermaneuverability, in broad terms, refers to the complex forms of non-conventional maneuverability that are found in high-performance combat aircraft. This capability includes maneuvers such as the Pugachev cobra, Kulbit and Herbst maneuver [1–3]; as well as broader, competing, classifications of flight behavior, including rapid nose-pointing-and-shooting (RaNPAS), pure sideslip maneuvering (PSM) [4,5] and direct force maneuvering (DFM) [6]. The development of supermaneuverable aircraft has been founded on advances in the study of unstable airframes, and the development of vectored propulsion technology [1,2]. Modern supermaneuverable aircraft remain characterized by these mechanisms; but increasing interdisciplinary contact with biological studies of maneuverability in flying creatures has led to parallel studies of an alternative, biomimetic, mechanism of supermaneuverability: one based on controlled wing morphing and motion. Thus far, biomimetic perching in unmanned aerial vehicles (UAVs) has been a central focus of these studies [7–9], with extensions into hover-to-cruise transition maneuvers [10], and incidence-based stall turns [11]. These maneuvers are primarily bio-inspired, and as such, studies of the biomimetic mechanism supermaneuverability have remained disjointed from studies of the thrust-vectored mechanism: the relationships between biomimetic and thrust-vectored maneuvers, mechanisms, and capabilities are rarely recognized [3].

To elucidate these relationships, we focus on a defined form of conventional supermaneuverability. Gal-Or [4,5] characterized rapid-nose-pointing-and-shooting (RaNPAS) capability as the control of aircraft orientation independent of its flight path, with specific reference to the orientation of fuselage-mounted weaponry. In the terminology of Herbst [6], this is a subset of direct force (DFM) capability, which includes also the capability for flight path control independent of orientation, rapid or otherwise. We consider the general case of nose-pointing-and-shooting (NPAS) capability, encompassing path-independent orientation control at all levels of transience. There are two key reasons to begin with the study of NPAS capability. (**a**) It delivers a broad advantage in aerial combat against non-supermaneuverable opponents, as per combat theory and simulation [4]. (**b**) Strongly-transient post-stall control is not necessarily required: even DFM capability at slower timescales is well beyond the capability of conventional aircraft. We refer to low-transience NPAS as quasistatic NPAS (QNPAS); and this will represent a particular focus of this study. The distinction between slow and rapid NPAS highlights a key, unrecognized, relationship between thrust-vectoring and biomimetic mechanisms of supermaneuverability: NPAS via thrust-vectoring may be costly, both in terms of fuel consumption and airspeed loss [1,4], and holding a nose orientation indefinitely is not feasible. Biomimetic mechanisms, relying on passive aerodynamic effects, may show greater potential for consistent longer-timescale operation.

Flight simulation is a useful context in which to study the maneuverability of biomimetic UAVs, and provide design guidance for the development of these systems. We approach the study of biomimetic NPAS/QNPAS through the simulation of a case-study system: a hybrid UAV system on the scale of ~1 m, equipped with a conventional propulsion system and, at maximum capability, six degree of freedom (6-DOF) wing morphing (asymmetric sweep, dihedral and incidence). This system is tailored to match the scale of a number of source aircraft and biological creatures, including the NextGen MFX-1 morphing-wing UAV [12], the remote-control ShowTime 50 [13], and greylag geese (*Anser anser*) [14–16]. To understand the QNPAS capability of this system, we develop a three-dimensional flight simulator for the system – using multibody analysis and an experimentally-validated section model – and utilize this to predict the aircraft's QNPAS flight envelope and open-loop response. We break new ground in the simulation of morphing-wing systems though the development of techniques for the analysis of multidimensional trim spaces. Through these novel techniques, we are able to identify complex phenomena in this class of biomimetic morphing-wing systems: the existence of, and key trends in, the multidimensional space of trim states; the mechanisms for generating and improving biomimetic QNPAS capability; and the interaction between QNPAS behavior and dynamic stability. We successfully demonstrate a wide range of QNPAS behavior. The capability of biomimetic systems for NPAS, in any form, has not previously been recognized: this study provides both the first concrete link between biological and thrust-vectored supermaneuverability; and a demonstration of the capability of even simple biomimetic systems for defined existing forms of supermaneuverability. This NPAS capability has implications for the design of artificially-intelligent UAV systems for dogfighting [17]: such systems may be able to extend their already impressive dogfighting ability into true supermaneuverability via biomimetic morphing-wing control.

## 2. Modelling

**2.1. Case study biomimetic morphing-wing UAV**

Case study UAV parameters are selected to match a range of source aircraft and biological creatures; Table 1 presents a parameter comparison. Figure 1 presents a scale rendering of the case study UAV, with active morphing degrees of freedom indicated, and a hypothetical air-combat NPAS maneuver schematic (cf. Gal-Or [4]) for reference. Note that the system wing mass is large (total 25% of the body mass), to account conservatively both for multi-degree-of-freedom actuation, and for wing strengthening to allow high-incidence states. We also intentionally decrease lifting surface chord values relative to their sources, so as to generate an aerodynamically suboptimal system that will represent a conservative estimate of potential system performance. The use of 6-DOF morphing represents a maximally-actuated configuration for the purposes of our scoping analysis: industrial applications would be expected to show a more restricted actuation space.

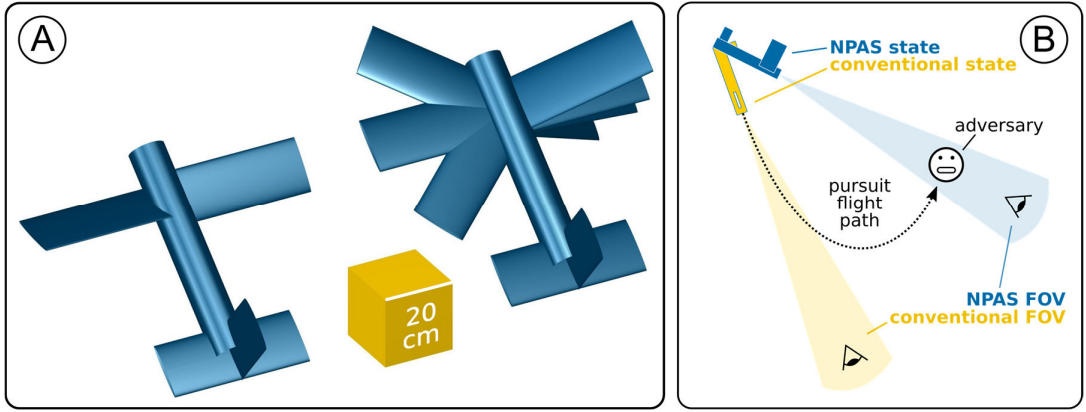

**Figure 1:** Case study biomimetic morphing-wing UAV with NPAS context. (**A**) Rendering of the case study UAV within the flight simulation code, illustrating the six morphing degrees of freedom: incidence (left) and sweep and dihedral (right). (**B**) Schematic, the vertical plane, of a hypothetical air-combat NPAS maneuver enabling fuselage-mounted field-of-view (FOV) contact with an adversary, cf. Gal-Or [4].

**Table 1:** Case study UAV properties: '/' denotes data not available or not applicable.

|  | This study | NextGen MFX-1 [12] | ShowTime 50 [13] | Greylag Goose (*A. Anser*) [14–16] |
|---|---|---|---|---|
| **Properties:** | **Values:** | | | |
| Length – fuselage | 1.20 m | 2.1 m | 1.51 m | ~0.8 m |
| Length – wing to tail | 0.80 m | 1.17 m | 0.94 m | / |
| Radius – body | 0.10 m | 0.15 m | 0.09 m | / |
| Span – wing | 1.60 m | 2.8 m | 1.46 m | ~1.6 m |
| Span – horz. stabilizer | 0.80 m | / | 0.62 m | / |
| Span – vert. stabilizer | 0.40 m | / | 0.17 m | / |
| Mean chord – wing | 0.15 m | 0.23 | 0.32 m | ~0.26 m |
| Mean chord – horz. stabilizer | 0.15 m | / | 0.22 m | / |
| Mean chord – vert. stabilizer | 0.15 m | / | 0.09 m | / |
| Airfoils | ST50 | / | ST50 | biological |
| Mass – total | 8 kg | 45 kg | 2.9 kg | ~3.3 kg |
| Mass – single wing | 1 kg | / | / | / |
| Propulsion – max. thrust | / | ~200 N | ~60N | / |
| Propulsion – mechanism | / | jet engine | propeller | flapping-wing |

## 2.2. Modelling context

For morphing-wing systems on the scale of ~1 m, a wide variety of structural-mechanic and aerodynamic modules are available as flight simulation components. Structural modules attested in morphing-wing literature range from Newtonian rigid-body models with time-dependent [18], or even time-independent [11], rotational inertia tensors; to exact multi-body formulations of the system's rigid [19] or continuum [20] mechanics. Newtonian rigid-body models typically neglect center-of-mass motion and momentum changes induced by morphing, but may include inertia tensor changes – an approximation typically made where only low levels of morphing deformation are required; for instance, in flight control [21–23]. In our case study UAV, we utilize a multibody dynamic model, to account for large wing masses and the potential for large-amplitude wing motion.

Aerodynamic modules available for morphing-wing systems show a similarly wide range of modelling fidelity. Three-dimensional turbulent computational fluid dynamics (CFD) simulations are occasionally utilized, typically an Reynolds-averaged (RANS/URANS) framework with a Spalart-Allmaras turbulence model [11,24]. Phenomenological dynamic stall and lift hysteresis models, such as the Goman-Khrabrov (GK) [25] model, available in a section-model context, are occasionally applied; and form the basis for simulations of biomimetic perching [9,26]. These models focus on the dynamic effects of airfoil pitching: the effect of dynamic sweep motion is only rarely studied [27], though results from the study of unsteady freestream flows indicate that it may have significance [28–30]. At a simpler level, quasisteady or steady section models or panel methods are available – including vortex-lattice [31,32], doublet-source [22] and lifting-line methods [26]. Second-order extensions to quasisteady section models are also available [33]. Panel methods have the advantage of including limited three-dimensional (e.g. finite-span) effects; but the disadvantage of being unable to be generalized to model dynamic stall or other unsteady effects. Most implementations are constrained to linear pre-stall aerodynamic models, and a generalization to even static stall behavior requires nonlinear lifting-line theory [34] or iterative decambering [35,36]. To enable a generalization to a dynamic stall modeling in further studies of RaNPAS capability, we begin here with a quasisteady section model, utilizing empirical data for aerodynamic coefficients over the entire angle of attack range, and accounting for flow induced by biomimetic wing motion.

### 2.3. Multibody dynamics

In the multibody-dynamic description of the case study UAV, the airframe is taken to consist of an isotropic cylindrical fuselage, a point mass for balancing purposes, vertical and horizontal stabilizers, and two movable wings of elliptical cross-section for inertial purposes. The orientation of the fuselage is parameterized with 3-2-1 Euler angles (vector $\boldsymbol{\theta}$), and its position by the location of a reference point ($S$) in earth reference frame; $\mathbf{x}_S^{(e)}$. Point $S$ the rearmost point on the fuselage, and the location stabilizer connections. The orientation vector $\boldsymbol{\theta}$ defines a rotation matrix, $R_{E/B}$, from earth ($E, e$) to body-fixed ($B, b$) reference frames [37]. The motions of the two wings relative to the fuselage are parameterized with left- and right-handed 3-2-1 Euler angles (vectors $\boldsymbol{\tau}_{wl}$ and $\boldsymbol{\tau}_{wr}$, respectively). In our open-loop study they are considered to be prescribed control parameters under the assumption of a perfect actuator.

The position, velocity and angular velocity ($\mathbf{x}_i, \dot{\mathbf{x}}_i, \boldsymbol{\omega}_i$) of each element in the multibody system, resolved in the earth reference frame, may be expressed as:

$$\mathbf{x}_i^{(e)} = \begin{cases} \mathbf{x}_S^{(e)} + \mathrm{R}_{E/i}(\boldsymbol{\theta})\mathbf{L}_i & i \in \mathcal{F} \\ \mathbf{x}_S^{(e)} + \mathrm{R}_{E/B}(\boldsymbol{\theta})\mathbf{L}_H + \mathrm{R}_{E/i}(\boldsymbol{\theta})\mathbf{L}_i & i \in \mathcal{W}, \end{cases}$$

$$\dot{\mathbf{x}}_i^{(e)} = \begin{cases} \mathrm{R}_{E/B}(\boldsymbol{\theta})\big(\dot{\mathbf{x}}_S^{(b)} + \boldsymbol{\omega}_0^{(b)} \times \mathrm{R}_{B/i}\mathbf{L}_i\big) & i \in \mathcal{F} \\ \mathrm{R}_{E/B}(\boldsymbol{\theta})\big(\dot{\mathbf{x}}_S^{(b)} + \boldsymbol{\omega}_0^{(b)} \times \mathbf{L}_H + \boldsymbol{\omega}_i^{(b)} \times \mathrm{R}_{B/i}\mathbf{L}_i\big) & i \in \mathcal{W}, \end{cases} \qquad (1)$$

$$\boldsymbol{\omega}_i^{(e)} = \begin{cases} \mathrm{R}_{E/B}(\boldsymbol{\theta})\boldsymbol{\omega}_0^{(b)} & i \in \mathcal{F} \\ \mathrm{R}_{E/B}(\boldsymbol{\theta})\big(\boldsymbol{\omega}_0^{(b)} + \Omega_i^{(b)}\dot{\boldsymbol{\tau}}_i\big) & i \in \mathcal{W}, \end{cases}$$

where $\mathrm{R}_{j/i}$ are transformation matrices from frame $i$ to frame $j$. $\mathcal{F}$ is the set of fuselage elements/frames and $\mathcal{W}$ the set of morphing-wing elements/frames; $B$ is the body frame and $E$ the earth frame. $\mathbf{L}_i$ are vectors of lengths denoting the position of the center of mass of each body in its own local reference frame; relative to $S$ for $i \in \mathcal{F}$ or the wing root $H$ for $i \in \mathcal{W}$. Additionally, $\mathbf{L}_H$ denotes the position of the root $H$ relative to $S$. $\Omega_i^{(b)}$ is the transformation matrix from the wing Euler angle rates $\dot{\boldsymbol{\tau}}_i$ to the wing angular velocities $\boldsymbol{\omega}_i^{(b)}$; and $\Omega_0^{(b)}$ the transformation matrix from the body Euler angle rates $\dot{\boldsymbol{\theta}}$ to the body angular velocity $\boldsymbol{\omega}_0^{(b)}$. Through an analysis of the system according to the non-conservative form of Lagrange's equations, we may derive the aircraft equations of motion as a system of first-order nonlinear ordinary differential equations (ODEs):

$$B_1(\mathbf{z})\dot{\mathbf{z}} = \mathbf{f}(\mathbf{z}) + \mathbf{f}_0(\mathbf{z}) - B_0(\mathbf{z})\mathbf{z}, \qquad (2)$$

where $\mathbf{z} = \left[\dot{\mathbf{x}}_S, \dot{\boldsymbol{\theta}}, \mathbf{x}_S, \boldsymbol{\theta}\right]^T$ is the aircraft state vector; $B_1(\mathbf{z})$ and $B_0(\mathbf{z})$ are matrix coefficients associated with the system multibody dynamics; $\mathbf{f}_0(\mathbf{z})$ is a vector forcing term associated with morphing-induced inertial loads; and $\mathbf{f}(\mathbf{z})$ is a vector forcing term associated with external system loads – aerodynamic, gravitational, propulsive, etc. Further details, including coefficient definitions for Eq. 2, are given in the Supplemental Materials.

### 2.4. Aerodynamic model

The term $\mathbf{f}(\mathbf{z})$ in Eq. 2 includes aerodynamic, gravitational and propulsive loads. Six independent load vectors must be defined: $Q_{\mathbf{x},\mathrm{gr}}$ and $Q_{\boldsymbol{\theta},\mathrm{gr}}$, the pair of generalized forces (in $\mathbf{x}$ and $\boldsymbol{\theta}$) associated with gravitational force; and the pairs $Q_{\mathbf{x},\mathrm{th}}$-$Q_{\boldsymbol{\theta},\mathrm{th}}$ and $Q_{\mathbf{x},\mathrm{ae}}$-$Q_{\boldsymbol{\theta},\mathrm{ae}}$, analogously associated with propulsive ($\cdot_{\mathrm{th}}$) and aerodynamic ($\cdot_{\mathrm{ae}}$) effects. Models of gravitational and propulsive effects are easily defined, and are given in the Supplemental Materials. The aerodynamic model is more complex. Each multibody element $i$ is discretized into $N_i$ aerodynamic stations ($i \in \mathcal{S}$) along its centroid, in a generalized form of section or blade-element model, as used by Ananda and Selig [38,39], among others. Aerodynamic reference points are located at the quarter-chord point, and station global velocities are computed via kinematics of Eq. 1, transformed into the appropriate local aerodynamic reference frame ($b_k$, with axes along the long and short airfoil axes, and the wing span). This generates a local flow velocity vector, which, under polar decomposition yields a local station angle of attack $\phi_{\text{eff},k}$, and airspeed magnitude $V_k$ within the local airfoil section plane. The spanwise component of the local flow velocity vector is neglected under the assumption that its associated skin friction drag is negligible, and that there is no flow shadowing, interaction, or propulsion-induced flow – several of the limitations of this model.

With the aerodynamic kinematics established, a local station quasistatic aerodynamic model is defined, with lift force (**L**), drag force (**D**), and pitching moment (**M**) vectors given according to the quasistatic model:

$$\begin{aligned}
\mathbf{L}_k^{(e)} &= \rho V_k^2 b_{c,k} C_{\text{L}}(\phi_{\text{eff},k}) \hat{\mathbf{L}}_k^{(e)}, \\
\mathbf{D}_k^{(e)} &= \rho V_k^2 b_{c,k} C_{\text{D}}(\phi_{\text{eff},k}) \hat{\mathbf{D}}_k^{(e)}, \\
\mathbf{M}_k^{(e)} &= \rho V_k^2 b_{c,k}^2 C_{\text{M}}(\phi_{\text{eff},k}) \hat{\mathbf{M}}_k^{(e)},
\end{aligned} \quad (3)$$

where $b_{c,k}$ is the local semichord, $\rho$ a constant air density, $C_L(\phi_{\text{eff},k})$ (etc.) the local aerodynamic coefficient w.r.t. angle of attack profiles, and $\hat{\mathbf{L}}_k^{(e)}$ (etc.) the force unit vectors in the earth frame – drag collinear with station airspeed, moment collinear with span, and lift perpendicular to both. Aerodynamic forces and moments are then integrated across the system, and transformed to the total generalized aerodynamic force and moment, $Q_{\mathbf{x},\text{ae}}$ and $Q_{\boldsymbol{\theta},\text{ae}}$ e.g., via $\Omega_0^{(b)}$ for $Q_{\boldsymbol{\theta},\text{ae}}$. In the case of the cylindrical fuselage, only drag force is computed, according to an assumed cylindrical fuselage cross-section. Finally, for the force coefficient profiles, $C_{\text{L}}(\phi_{\text{eff},k})$, etc., we use data from Selig [39] for proprietary wing and stabilizer airfoils; generated via data fusion techniques from experimental results, potential flow models and semi-empirical techniques. This data covers the full $\alpha$ range and is parameterized with respect to aileron and stabilizer deflections.

This completes the definition of the quasisteady section model for the case study UAV. This model is valid for all lifting-surface angles of attack, and accounts for quasistatic stall and morphing-induced flow, but no other unsteady aerodynamic effects. For flight simulation, we utilise MATLAB's inbuilt adaptive RK4(5) solver, ODE45, applied to Eq. 2. To avoid gimbal lock, we implement a pole-switching routine, performing an Euler angle conversion (3-2-1 to 2-3-1, and vice versa) when the simulation draws near an Euler angle pole. This conversion occurs at the end of an integration step, preventing any finite-difference approximations from bridging a pole switch. The system multibody dynamic model is then verified via an independent quaternion formulation: the details are tangential to this paper; and subsidiary to the model validation, which follows.

## 3. Validation

We validate the complete flight dynamic model against a nonlinear stability-derivative model of a 0.4-scale RQ-2 Pioneer UAV. Wind-tunnel data for this aircraft, from Bray [40], has been implemented by Selig and Scott [41] into the architecture of open-source flight dynamic model *JSBSim* [42], within flight simulator *FlightGear* [43]. This yields a nonlinear flight dynamic model, valid for angle of attack range $-7°$ to $17°$. Both the wind-tunnel data behind this model, and the model's flight simulation results, can be used to validate our model. First, to simulate the RQ-2 with our model, the airframe configuration is altered to have an H-tail and a shoulder wing. Airfoil aerodynamic data is taken from Critzos et al. [44] (NACA0012) and Ostowari and Naik [45] (NACA4415). Body drag coefficients are identified from data in Bray [40]. Geometric data is taken from Bray [40], as well as scale drawings [46], though here the UAV model number (RQ-2B/RQ-2A) differs. The UAV mass distribution is matched to two sources: the known inertia tensor of the aircraft [41], and a specified center-of-mass location, not directly known.

As a first validation, we compare nonlinear aerodynamic coefficient functions derived from data in Bray [40] with predictions from our model – as per Figure 2. Overall, the results are good: trends in all aerodynamic coefficients are predicted accurately. The stall transition in our model is earlier than indicated in the experimental data; probably as result of difference in airfoil stall point between our aerodynamic data and the conditions in Bray [40]. As a second validation, we compare flight simulation predictions between the *JSBSim* model and our model – a comparison that includes the effects of both linear and nonlinear stability derivatives. Figure 3 shows the orientation histories for six validation maneuvers (A-F), corresponding to three forms of aircraft behavior – phugoid oscillations (A-B), dives (C-D) and spiral dives, with large and small initial bank (E, 37° and F, 17°). Results for two center-of-mass locations are presented: one fitted to pitch-dominant simulations (phugoid and dives); one to yaw and roll-dominant simulations (spiral dives). Again, the results are good. Dive simulations are robust with respect to center-of-mass location; and phugoid simulations are well-modelled by the pitch-matched location; with roll and yaw destabilization captured. The pitch-matched location also generated good results for the large-perturbation spiral dive; only losing precision in the small-perturbation spiral dive. Overall, our flight dynamic model is validated, in a regime of low aerodynamic unsteadiness.

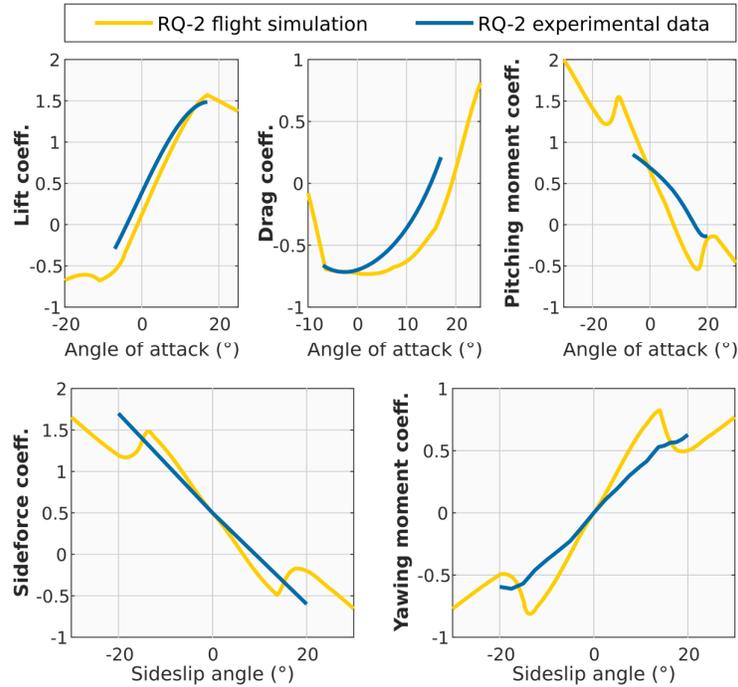

**Figure 2:** Comparison of experimental nonlinear coefficients from Bray [40], for the RQ-2 Pioneer UAV, with the predictions of our simulation: coefficients of lift, drag, sideforce, pitching moment and yawing moment.

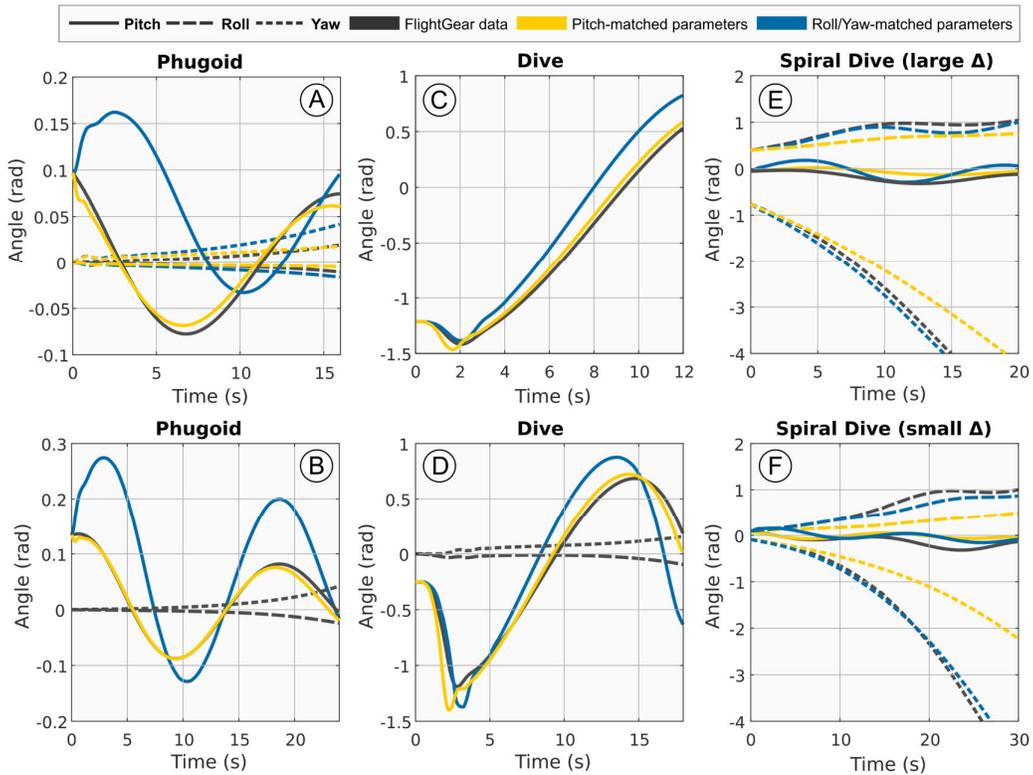

**Figure 3:** Simulation validation results for the RQ-2 Pioneer UAV model: orientation histories under six maneuvers, A-F. These maneuvers correspond to three forms of aircraft behavior – phugoid oscillations (A-B), dives (C-D) and spiral dives, with large and small initial bank perturbation (E, 37° and F, 17°). Note that orientation histories not visibly distinct from zero are omitted for clarity.

# 4. Trim state analysis

Even in the case of steady level flight, at a given airspeed, a key difference exists between conventional aircraft and the case study UAV and: in the former, there exists only a single trim state; but in the latter, there exists a multidimensional space of trim states. In our flight dynamic model, trim states are defined by the condition of zero translational and angular acceleration, that is:

$$\dot{\mathbf{z}}([U_{tr}, \boldsymbol{\theta}_{tr}], \mathbf{v}) = \dot{\mathbf{z}}(\mathbf{z}_{tr}, \mathbf{v}) = [0_{1\times6}, U_{tr}, 0_{1\times5}]^T,$$

$$\text{with } \dot{\mathbf{z}}(\mathbf{z}_{tr}, \mathbf{v}) = B_1(\mathbf{z}_{tr}, \mathbf{v})^{-1}(\mathbf{f}(\mathbf{z}_{tr}, \mathbf{v}) - B_0(\mathbf{z}_{tr}\mathbf{v})\mathbf{z}_{tr}),$$

$$\text{and } \mathbf{z}_{tr} = [U_{tr}, 0_{1\times8}, \boldsymbol{\theta}_{tr}]^T,$$

(4)

where $U_{tr}$ and $\boldsymbol{\theta}_{tr}$ are the trim state ($tr$) airspeed and orientation. If this airspeed and orientation are given, then we may solve Eq. 4 the set of control and/or structural parameters, $\mathbf{v}$, that will trim this given state. In general, the 12-DOF condition on $\mathbf{v}$, $\dot{\mathbf{z}}(\mathbf{z}_{tr}, \mathbf{v}) = [0_{1\times6}, U_{tr}, 0_{1\times5}]^T$, reduced to a 6-DOF condition, $F(\mathbf{v}) = 0_{1\times6}$; but in the case of motion within a single plane, the condition may reduce further. For instance, in the case of pitching motion, $\boldsymbol{\theta}_{tr} = [\alpha_{tr}, 0, 0]^T$ for angle of attack $\alpha_{tr}$, the condition, Eq. 4, reduces to 3-DOF, $F(\mathbf{v}) = 0_{1\times3}$, with components of pitching and non-lateral translational acceleration. Appropriate control variables are transparent: elevator deflection, symmetric wing incidence, and thrust force; $\mathbf{v} \in \mathbb{R}^3$. We explore this control space via natural continuation [47]: using Newton's method to compute a solution at an initial value ($\alpha_{tr} = 0$, $U = 25$ m/s), we increment $\alpha_t$ and $U_t$ and utilize the existing solution as an initial guess for the solution of the incremented problem.

Figure 4 shows resulting trim space in angle of attack. Several points may be noted. For the case study UAV, the limits of the trim space are determined solely by the elevator limits (~0.87 rad [39]), and increases in elevator control effectiveness are an avenue to a wider trim space. Physically, obtaining a trim state at high angle of attack requires generating lift forces on the wings and horizontal stabilizer that are near-identical to their trimmed values at zero angle of attack: for the wing, by maintaining near-constant angle of attack; and for the horizontal stabilizer, via elevator deflection. The small nonlinearities in this relation account for (**a**) differences moment arm due to trigonometric effects of the vertical ($z$) position of the aircraft center of mass; and (**b**) system-wide drag effects. These factors point to innovative structural approaches to increase the trim space size – for instance, generating additional tailplane drag (e.g., via airbrakes).

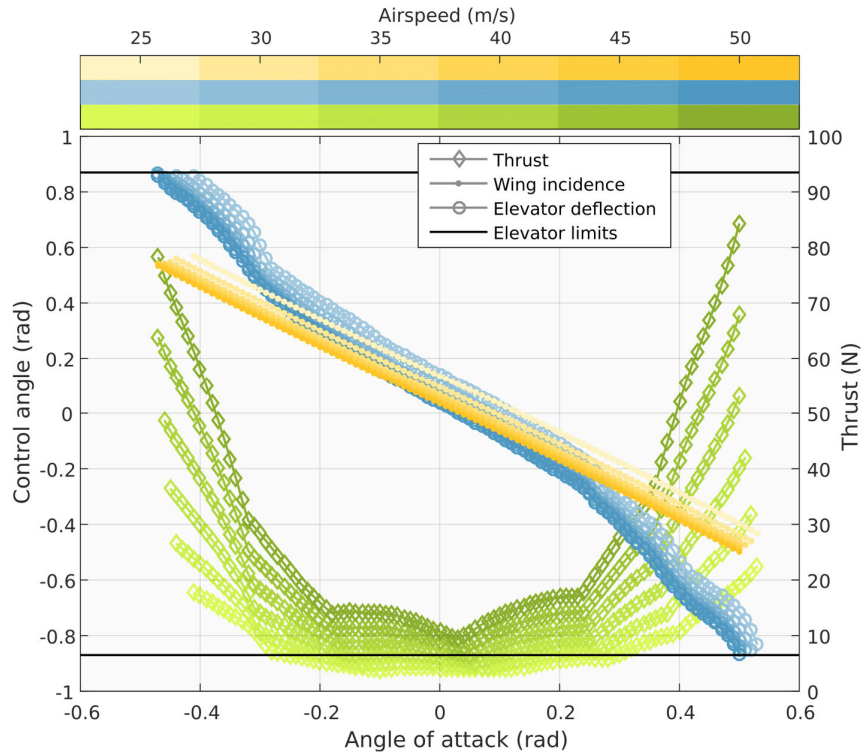

**Figure 4:** Case study UAV trim space in angle of attack, showing control parameters: thrust, symmetric wing incidence, and symmetric elevator deflection. As can be seen, the elevator limits are the primary factor limiting the size of the trim space.

The same principle, but at greater complexity, applies to the trim space of the yawing aircraft; $\boldsymbol{\theta}_{tr} = [0, \beta_{tr}, 0]^T$ for sideslip angle $\beta_{tr}$. Motion within the horizontal plane does not lead to control symmetry, and does not reduce the size of the 6-DOF condition – the angle of attack degree of freedom can be included in the trim state at no cost in complexity. However, this two-angle trim state problem, with $\boldsymbol{\theta}_{tr} = [\alpha_{tr}, \beta_{tr}, 0]^T$ and a 6-DOF residual, remains underconstrained: there are nine control variables. The thrust, elevator and rudder deflection, and asymmetric incidence are five control variables that would be unreasonable to remove, ensuring constrainedness involves reducing asymmetric sweep and dihedral control to a single degree of freedom. A simple way to do this is to select single-wing (right/left) control in sweep or dihedral, and constrain the non-selected wing to a defined value (initially, zero sweep/dihedral). The two-dimensional trim space of the aircraft can then be computed via natural continuation. Note that the size of the trim space only slightly dependent on airspeed, with <5% variation over airspeed 25 m/s to 50 m/s.

Figure 5 shows the trim space available under left-wing dihedral and sweep control. Reversing the yaw axis yields the equivalent result for right-wing control. In maneuver terms it is more insightful to look at this orientation in relative terms: inboard/outboard of the current yawed trim state. For instance, in a trim state of leftwards yaw, the right wing is inboard, and the left outboard. Figure 5 designates the morphing control in terms of inboard or outboard constraints (non-actuation). Comparing dihedral control to sweep control, in terms of trim space size, we see that dihedral control is superior: sweep control shows poorer performance at low angles and identical maximal performance in some orientation sectors. The latter implies that some additional parameter is the limiting factor on trim space size in these sectors. This can be confirmed: the elevator and rudder reach limit states on the entire boundary of the dihedral-morphing trim space. Notably, the elevator and rudder alone cannot generate a trim space over sideslip angle and angle of attack; yet in this system they are the controlling factor in the size of this trim space. As a further explanation of the mechanism of yawed state trimming, Figure 6 shows a visualization of the system trim space under dihedral control, with the space of available fuselage axes alongside four full-system test states. As can be seen, dihedral angle (of either handedness) is used to generate an inboard stabilizing lift force; countering fuselage drag and vertical stabilizer effects that generate an outboard force.

The existence of the wide and accessible trim space shown in Figures 5-6 is crucial to NPAS capability in this system. This space control allows level flight – stable level flight if the trim state is stable – at any point in the trim space: up to angle of attack 30° and sideslip 20° in the case study UAV. By inference; if these trim states are stable or can be stabilized, then quasistatic NPAS (QNPAS) orientation capability is available throughout the orientation range of the trim space; simply by changing the aircraft configuration through the space of trim configurations. There is, however, one further parameter to consider: the value of the dihedral constraint itself, which can be nonzero. This introduces an additional parameter into the trim space dynamics: $\Gamma$, the dihedral constraint. Defining a nonzero constraint on the non-morphing wing is equivalent to defining the morphing control motions as deviations from an analogous symmetric-dihedral state: a state with wings at 0° and 20° dihedral represents both a single-wing 20° deviation from symmetric 0° dihedral; and a single-wing −20° deviation from symmetric 20° dihedral. This distinction highlights the way in which $\Gamma$ may have a significant effect on the flight dynamic stability of the system trim states. This may be confirmed: analysis up to $\Gamma = 0.4$ rad (20°) indicates that, over this range, there is minimal change in trim space size and shape; but significant changes to the flight dynamic stability of the trim states, as we demonstrate in Section 5.

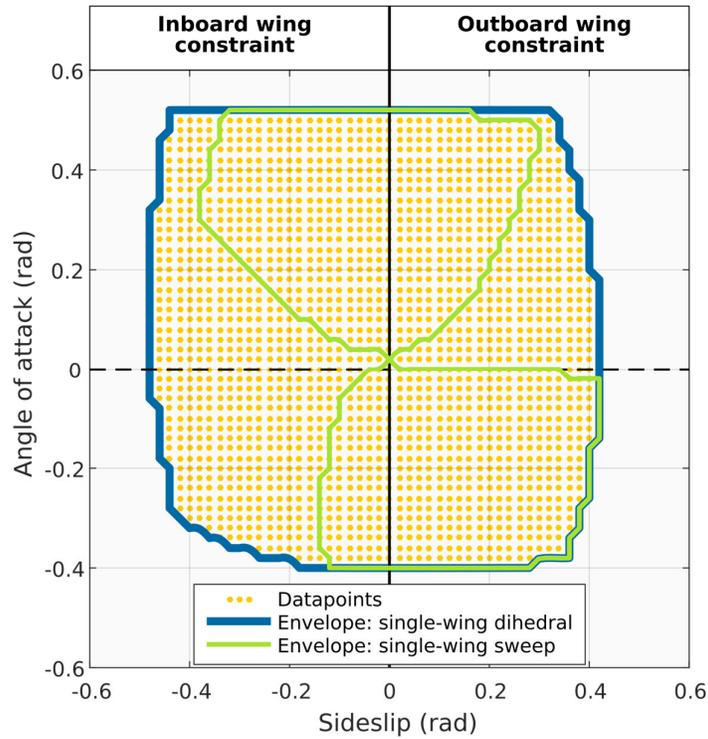

**Figure 5:** Case study UAV trim space in combined angle of attack and sideslip, under single-wing dihedral and sweep morphing, airspeed 25 m/s. In the case study UAV, the sweep-morphing trim space is contained wholly within the dihedral-morphing trim space.

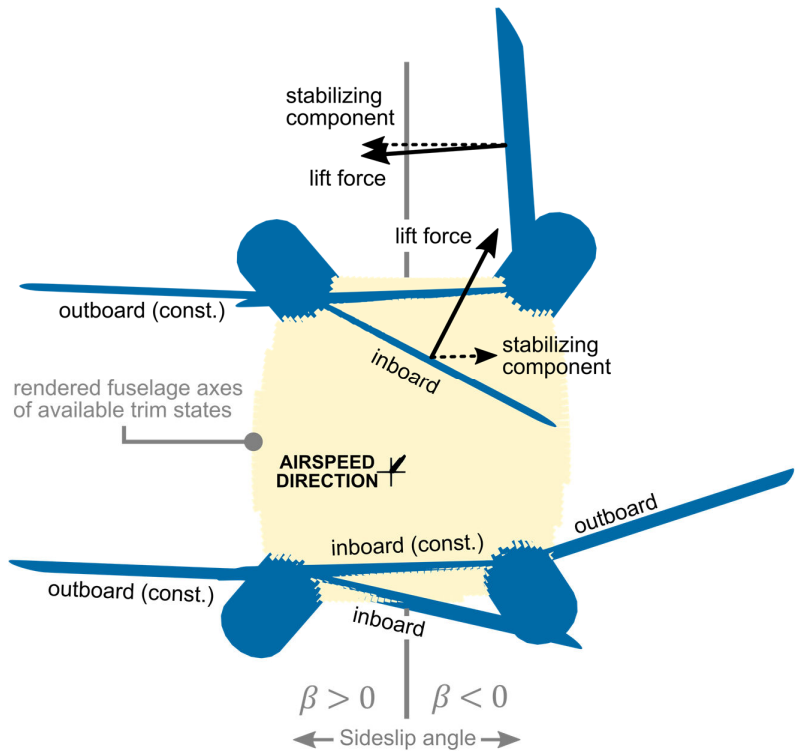

**Figure 6:** Rendering of the case study UAV trim space under single-wing dihedral control: fuselage axes of all available trim states, with full system renderings for four extreme test states. The distinction between inboard and outboard dihedral constraints is shown; and a lateral trimming mechanism is indicated.

# 5. QNPAS flight-dynamic stability

## 5.1. Longitudinal and lateral stability analysis

The stability of states in the trim space of the case study UAV is of key relevance to its QNPAS capability. Static stability across the trim space can be characterized via longitudinal and lateral acceleration gradients $\partial\ddot{\alpha}/\partial\alpha$ and $\partial\ddot{\beta}/\partial\beta$, where $\ddot{\alpha}/\ddot{\beta}$ represent the longitudinal/lateral acceleration outputs of Eq. 2 (within $\dot{\mathbf{z}}$). An analysis of these metrics (not presented) indicates broad longitudinal and lateral stability across the case study UAV, with the exception of a tiny zone of lateral instability at high angle of attack under positive-dihedral ($\Gamma = 0.3$ rad) inboard constraint. The overall picture is one of consistent static stability, indicating broad utility for these asymmetric trim states without static stability augmentation. The contrast with thrust-vectoring supermaneuverable aircraft, utilizing unstable airframes, is also notable.

## 5.2. Dynamic and spiral stability analysis

The dynamic stability properties of a trim point can be computed by linearizing the first-order nonlinear system. Eq. 2, governing the system dynamics:

$$\dot{\mathbf{z}} = \mathbf{S}(\mathbf{z}, \mathbf{v}) \cong \mathbf{S}(\mathbf{z}_{tr}, \mathbf{v}) + \left.\frac{\partial \mathbf{S}(\mathbf{z}, \mathbf{v})}{\partial \mathbf{z}}\right|_{\mathbf{z}=\mathbf{z}_{tr}} (\mathbf{z} - \mathbf{z}_{tr}), \qquad (5)$$

where $\mathbf{z}_{tr}$ is the local trim state, and the system function $\mathbf{S}(\mathbf{z}, \mathbf{v})$ may be represented:

$$\mathbf{S}(\mathbf{z}, \mathbf{v}) = B_1(\mathbf{z}, \mathbf{v})^{-1}(\mathbf{f}(\mathbf{z}, \mathbf{v}) - B_0(\mathbf{z}, \mathbf{v})\mathbf{z}). \qquad (6)$$

The eigenvalues of the Jacobian $\partial \mathbf{S}(\mathbf{z}, \mathbf{v})/\partial \mathbf{z}|_{\mathbf{z}=\mathbf{z}_{tr}}$ then yield the flight dynamic modes of the system, in terms of the local orientation coordinates $(\mathbf{z} - \mathbf{z}_{tr})$. These modes can be transformed to orientation coordinates defined around the velocity vector, as per the conventional description of flight dynamic modes. While the system stability space is complex, there are only two key forms of dynamic instability. The first is a Dutch roll instability in the positive-dihedral system ($\Gamma = 0.3$ rad) at extreme trim state pitch angles. Figure 7 shows this stability zone over the system trim space; with representative modal pathways, and flight simulation results for validation. The key implication of this instability is to render open-loop QNPAS control unavailable in unstable areas of the trim space – particularly, $\alpha_{tr} < 0$ for $\Gamma = 0.3$ rad.

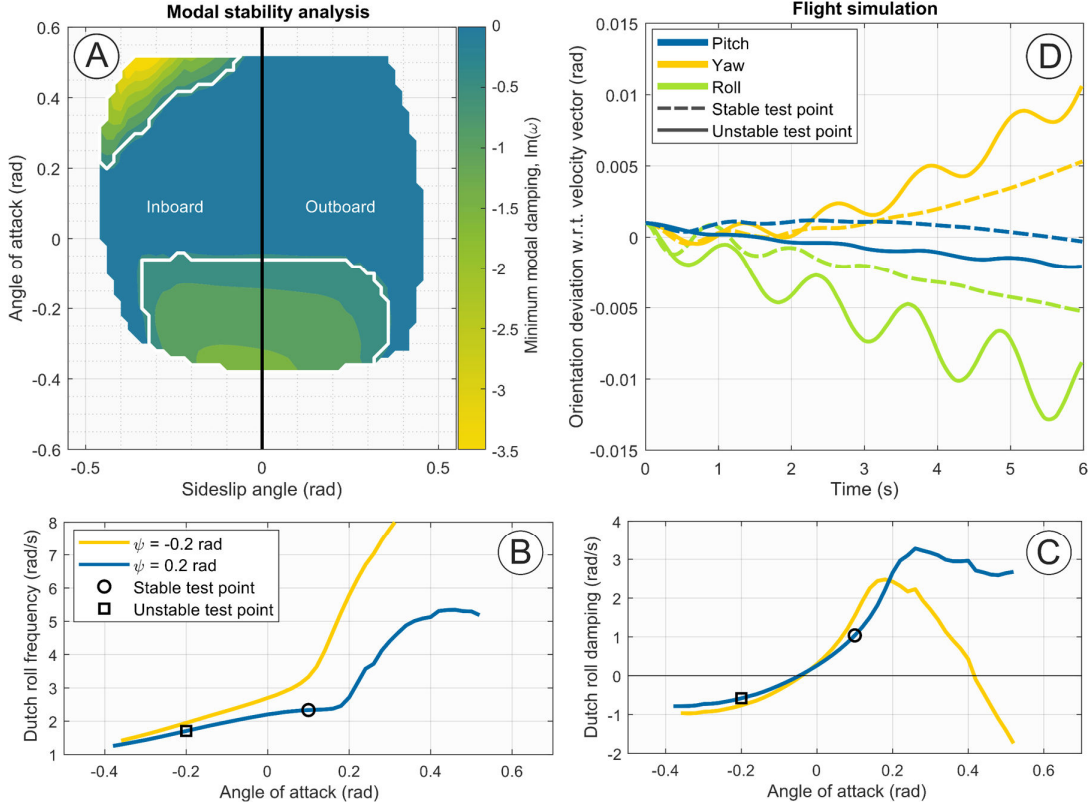

**Figure 7:** Modal dynamics of Dutch roll instability in the case study UAV ($\Gamma = 0.3$ rad). **A**: Trim space modal damping predictions. **B-C**: representative modal frequency and damping paths w.r.t. angle of attack. **D**: flight simulation validation for stable/unstable test points.

The second form of dynamic instability is a widespread but complex set of instabilities related to the spiral and roll subsidence modes. The modal pathways are difficult to disentangle; so, to get an overview of the impact of this instability on the system, we perform numerical perturbation on areas unaffected by Dutch roll instability. We simulate the system response to yaw perturbation ($\Delta\psi = 0.05$ rad), and evaluate yaw and roll deviation metrics ($\delta_\psi$, $\delta_\phi$ at the simulation endpoint ($t = t_{\text{end}} = 15$ s):

$$\delta_\psi = |\psi|_{t=t_{\text{end}}} - \psi_{tr}|/\Delta\psi, \qquad \delta_\phi = |\phi|_{t=t_{\text{end}}}|/\Delta\psi. \tag{7}$$

The results of these simulations align with classical results on the effect of dihedral on dynamic stability [48] – details are presented in the Supplemental Materials. Increased biomimetic dihedral constraint, representing increased average dihedral in the system, is associated with decreased spiral mode-type instability. The difference is notable. The positive-dihedral system ($\Gamma = 0.3$ rad) shows typical metrics $0.01 \leq \delta_\phi \leq 0.1$ and $0.2 \leq \delta_\psi \leq 1$; and the zero-dihedral system ($\Gamma = 0$) typical metrics $0.1 \leq \delta_\phi \leq 1$ and $0.5 \leq \delta_\psi \leq 3$. Notably, at zero dihedral, the inboard-constrained system shows significantly weaker spiral instability than the outboard-constrained

Page **17** of 32

system: this alludes to the need for the outboard-constrained system to use significant anhedral on the inboard wing to generate a lateral stabilizing force – see Figure 6. The same need explains the strength of Dutch roll instability under the inboard constraint: a decrease in the average dihedral of the system decreases the Dutch roll stability [48]. Choices of dihedral constraint (value and type) have a significant impact on trim space flight dynamic stability – and the effect of this stability on QNPAS behavior is complex, as we illustrate in Section 6.

## 6. QNPAS guidance and simulation

### 6.1. QNPAS continuation design

At this point, all the contextual information required to simulate biomimetic QNPAS behavior is available. A QNPAS maneuver is designed using the following principles. The target motion is defined by time functions in angle of attack $\alpha_{tg}(t)$, sideslip angle $\beta_{tg}(t)$, roll angle $\gamma_{tg}(t)$, and airspeed $U_{tg}(t)$. An initial condition is selected for the start time ($t = 0$): typically, $\alpha_{tg}(0) = \beta_{tg}(0) = 0$. A trim state at this initial condition is computed by solving the nonlinear trim state system, Eq. 4, via Newton's method, with $\alpha_{tr} = \alpha_{tg}(0)$, $\beta_{tr} = \beta_{tg}(0)$, and $U_{tr} = U_{tg}(0)$. If the target functions $\alpha_{tg}(t)$ and $\beta_{tg}(t)$ are continuous, then the associated trim state control configurations can be computed via natural continuation directly along the target pathway – in the same way that the trim space was initially explored.

That is, utilizing the nomenclature of Eq. 4, the trim state at discrete timestep $t_k$ is computed as the solution of the nonlinear trim residual function in terms of $\mathbf{v}_k$, as a function of the target profile:

$$\text{Trim residual function:} \quad \dot{\mathbf{z}}(\mathbf{z}_{tr,(k)}, \mathbf{v}_k) = [0_{1\times 6}, U_{tg}(t_k), 0_{1\times 5}]^T,$$

$$\text{Target trim state vector:} \quad \mathbf{z}_{tr,(k)} = [U_{tg}(t_k), 0_{1\times 8}, \boldsymbol{\theta}_{tr,(k)}]^T, \qquad (8)$$

$$\text{Target Euler angle vector:} \quad \boldsymbol{\theta}_{tr,(k)} = [\alpha_{tg}(t_k), \beta_{tg}(t_k), \gamma_{tg}(t_k)]^T,$$

where $\mathbf{v}_k$ is a vector of configuration parameters (morphing and control surface deflections), directed to appropriate aerodynamic/multibody model contexts. The definition of $\mathbf{v}_k$ is particularly complex in the case of inboard/outboard switching motion (Section 4); in which the right/left wing dihedral are actuated only when they are inboard or outboard relative to the trim state. In this scenario, a single proxy dihedral parameter from the Newton iteration is directed to the left or right wing depending on the local target (*not* observed) trim state sideslip value. The initial guess for the Newton iteration for $\mathbf{v}_k$ is the configuration $\mathbf{v}_{k-1}$ (natural continuation [47]).

Continuous-time solutions for $\mathbf{v}_k$ may be available for analytical formulations of $\dot{\mathbf{z}}(\cdot)$, but this is beyond the scope of this study.

In this way, an efficient and generalizable method of obtaining QNPAS commands is obtained. Here we study a slightly simplified case, with constant airspeed, $U_{tg}(t) = U_c = 25$ m/s; and target zero roll, $\gamma_{tg}(t) = 0$. Changes in airspeed affects only the required propulsive thrust values, with minimal impact on trim state control surface values (Figure 4); and roll control in an NPAS context is not notable: conventional aircraft are capable of this. For initial studies, we utilize a scroll-shaped maneuver form defined by:

$$r = \begin{cases} 0.5(1 - \cos(2\pi t/T)) & 0 \leq t \leq T/2 \\ 1 & t \geq T/2, \end{cases}$$
$$\alpha_{tg}(t) = r(\hat{\alpha}\cos(2\pi t/T) + \alpha_{cp}), \qquad (9)$$
$$\beta_{tg}(t) = r(\hat{\beta}\sin(2\pi t/T) + \beta_{cp}),$$

with oscillatory period $T$, angle of attack/sideslip angle amplitudes $\hat{\alpha}$ and $\hat{\beta}$; and angle of attack/sideslip angle center-points $\alpha_{cp}$ and $\beta_{cp}$. Figure 8 shows two representative histories for this scroll-shaped path.

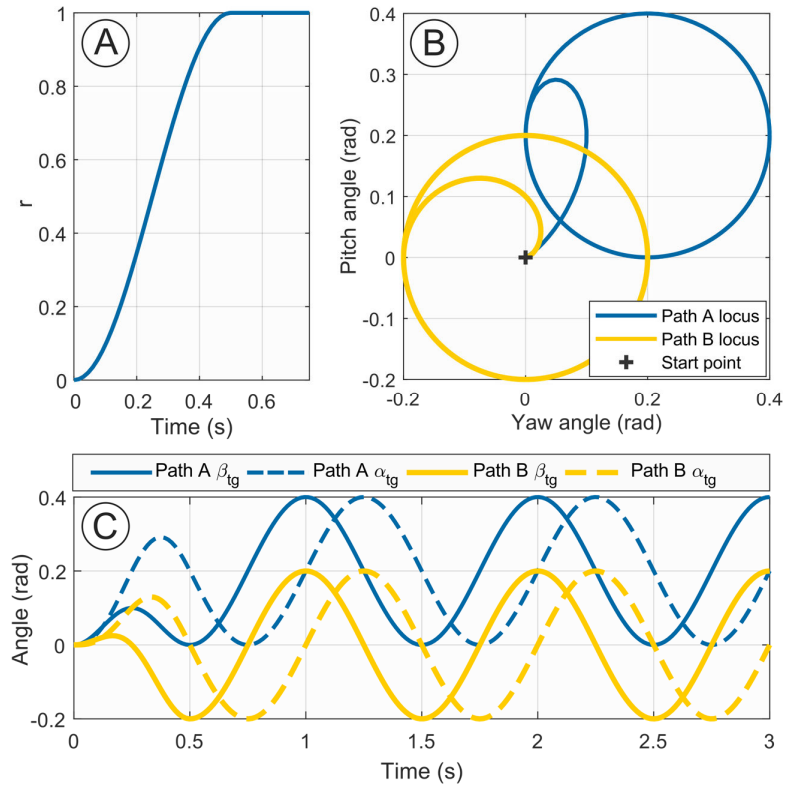

**Figure 8:** Two representative scroll-shaped paths. **A**: Ramp function $r(t)$. **B**: Target path loci in pitch and yaw (angle of attack/sideslip angle). **C**: Time histories in pitch and yaw. Parameter values are: Path A: $\alpha_{cp} = \beta_{cp} = 0.2$ rad; Path B: $\alpha_{cp} = \beta_{cp} = 0$ rad; both with $\hat{\alpha} = \hat{\beta} = 0.2$ rad.

### 6.2. QNPAS tracking and transience thresholds

The trim state continuation method described in Section 6.1 is strictly valid only for quasistatic changes in the target orientation. Two classes of effect contribute to a breakdown in trim state tracking for transient changes in target orientation. The first are the effects of QNPAS-induced and morphing-induced flow: these induced flows alter the effective angles of attacks and airspeeds of the lifting surfaces. These effects are still quasisteady, in that they can be characterized by the instantaneous state of the aircraft (e.g., wing and body velocity). They are modelled in the flight simulation, but not included in the trim state target formulation. The second class of effects are the unsteady effects arising from QNPAS motion: at slower timescales, classical unsteady aerodynamics [49]; and at faster timescales, dynamic stall effects [25]. In the context of this study, the presence of these effects indicates that an NPAS maneuver should properly be considered rapid (RaNPAS), as per Gal-Or [4,5]; as opposed to quasistatic (QNPAS). These unsteady effects are not included in the flight dynamic model – instead, we will be concerned with establishing the limits within which these effects will not be significant.

Regarding the impact of the quasisteady effects of QNPAS-induced flow: this can be tested directly, by applying the trim continuation method of Section 6.1 to the scroll-shaped orientation target in independent pitch and yaw planes, we test the precision of the QNPAS response. Figures 9 and 10 show the results for independent longitudinal (pitching/angle of attack) and lateral (yawing/sideslip angle) motion, for timescale $10 \leq T \leq 40$ s. The impact of unsteady aerodynamic effects can be estimated by characterizing QNPAS maneuvers in terms of the lifting surface reduced frequencies, which can then be correlated to the expected significance of local unsteady effects. Reduced frequency, $\kappa_k = b_k \omega_k / V_k$ for aerodynamic station $k$, is a spectral quantity, based on the angular frequency $\omega_k$ of the local angle of attack history $\phi_k(t)$; local airspeed $V_k$; and local semichord $b_k$. When $\phi_k(t)$ is not simple-harmonic, its reduced frequency spectrum, $\Phi_k(\kappa_k)$, can be computed as a transformation of its angular frequency spectrum, $\Phi_k(\omega_k)$, the latter computed by Fourier analysis. In the case study UAV, probes ($k = m$) are taken on the wing and stabilizer tips – confirmed to be the location of greatest angle of attack variation – and the amplitude spectra of the probes $\phi_m(t)$ are computed; $|\Phi_m(\omega_m)|$. Through $\kappa_m = b_m \omega_m / V_m$, this is transformed to $|\Phi_m(\kappa_m)|$, but with a catch: the airspeed $V_m(t)$ is slightly time-dependent. In the absence of any established method for computing the reduced frequency in this context, a simple bounding method retains accuracy: utilizing $\min V_m(t)$ or $\max V_m(t)$ yields a maximal or minimal (respectively) bound on the local $\kappa_k$ value: in between is the band in which the exact $\kappa_m$ is guaranteed to lie.

It is commonly estimated [50–53] that the zone $0.01 < \kappa < 0.05$ represents the threshold between quasistatic and unsteady aerodynamic behavior. We take $\kappa = 0.01$ as a conservative threshold value: this threshold is equivalent to a critical timescale, $t_k^* = 2\pi/\omega_k$. Figures 9-10 show histories and power spectra for the effective angle of attack of each relevant lifting surface tip (accounting for near-symmetry), in the case of the most rapid longitudinal and lateral maneuvers ($T = 10$ s), indicating both the $\kappa = 0.01$ threshold value, and the target forcing frequency ($\propto 1/T$). Several points may be noted. (**a**) Overall, the stabilizers are subject to the greatest unsteadiness – incidence morphing ensures that the wing remains within a limited, pre-stall, angle of attack range throughout. (**b**) In the longitudinal maneuver, the angle of attack spectral content above the unsteadiness threshold is negligible. Quantitatively, in horizontal stabilizer, the peak spectral value above the unsteadiness threshold is ~1/50,000 the power of the target frequency ($T$) peak; and the peak angle of attack amplitude above the threshold is 0.05°. For the wings, the analogous power factor is ~1/750, and the peak angle of attack amplitude, 0.01°. (**c**) The lateral maneuver, while showing more unsteadiness, still remains within the zone of expected quasisteady behavior: across all lifting surfaces, the maximum angle of attack amplitude, in the unsteady zone, is 0.2° (on the vertical stabilizer) and the maximum power factor is ~1/260 (on the horizontal stabilizer).

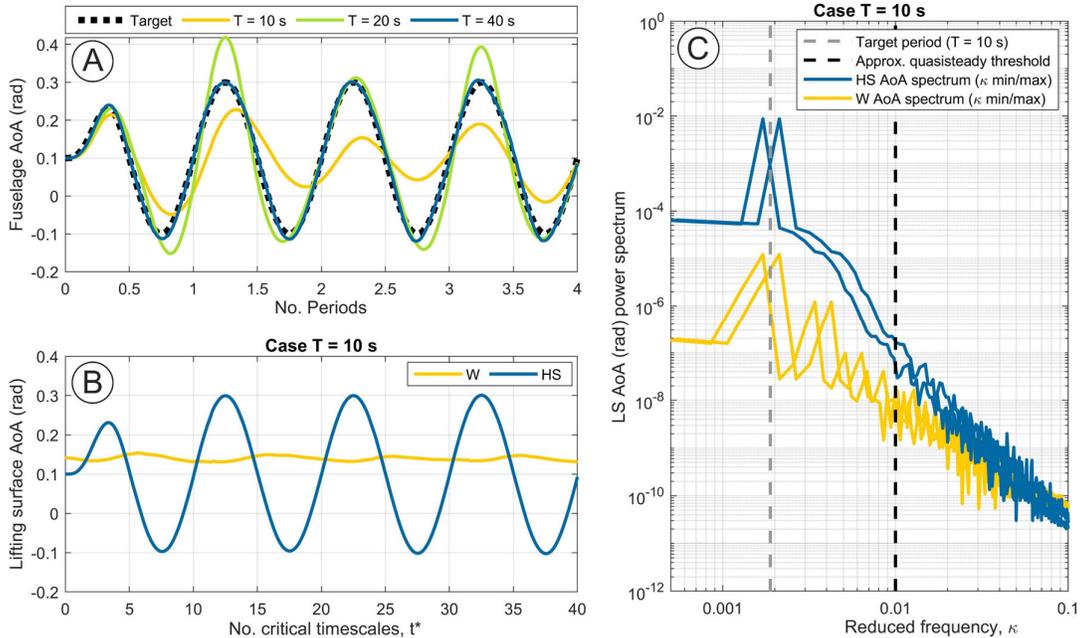

**Figure 9:** Pitch-axis QNPAS. **A**: Fuselage angle of attack tracking results w.r.t. the target period $T$. **B**: Lifting surface angle of attack histories; wing (W) and horizontal stabilizer (HS). **C**: Reduced frequency ($\kappa$) spectra of the lifting surface angles of attack, for minimum and maximum estimates of $\kappa$ (the dual curves), and indicating the QNPAS target period and an approximate threshold for quasisteady aerodynamic model validity.

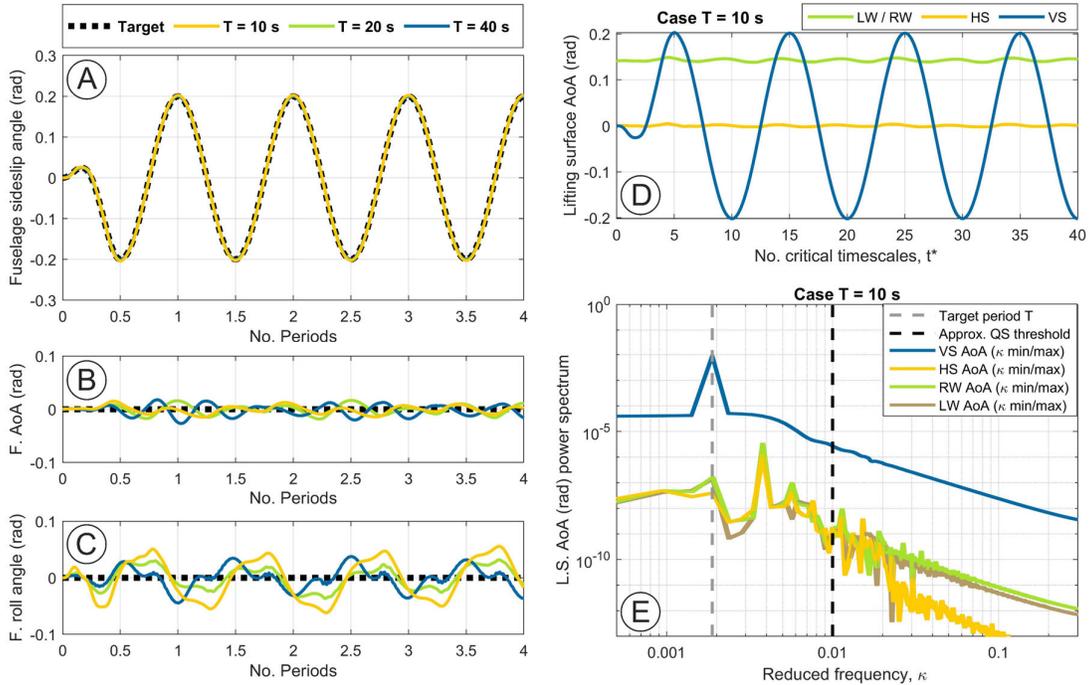

**Figure 10:** Yaw-axis QNPAS. **A**: Tracking results in the sideslip angle w.r.t. the target period $T$; **B**: in the angle of attack, target zero; **C**: in the roll angle, target zero. **D**: Lifting surface angle of attack histories; left and right wing (LW/RW), visually identical at displayed scale; horizontal stabilizer (HS) and vertical stabilizer (VS). **E**: Reduced frequency ($\kappa$) spectra of the lifting surface angles of attack, for minimum and maximum estimates of $\kappa$, with dashed lines indicating the QNPAS target period and an approximate threshold for quasisteady aerodynamic model validity.

With quasisteady behavior established, we may now assess the impact of quasisteady effects vis-à-vis the quasistatic QNPAS control: imprecisions observed in the target tracking are necessarily due to quasisteady induced-flow effects. A key observation is that lateral tracking is significantly more precise than longitudinal tracking: precise sideslip angle tracking is achieved for $T = 10$ s; but analogously, precise angle of attack tracking is not achieved even by $T = 40$ s. However, imprecision in lateral tracking tends to be expressed not in the sideslip angle; but in the aircraft roll angle and angle of attack. Physically, these observations are consistent: induced flow from airframe pitching motion will alter airfoil effective angles of attack, impacting pitch trim state behavior; whereas yawing motion will induce lateral gradients in effective airspeed, leading to large roll deflections which do not directly impact the yaw trim state. These roll motions can be seen in Figure 10: the yaw trim state is remarkably robust, retaining high precision even up to ~30° roll. A query may be raised regarding the possibility of even faster yawing timescales, given the precision of the yaw response: these are possible, but their accuracy under quasisteady aerodynamics becomes uncertain. In Section 7, a faster-timescale response using pitch-yaw entrainment is subject to a preliminary analysis.

# 7. Complex QNPAS

## 7.1. Multiaxis QNPAS

Having established the capability of the system for independent-axes QNPAS, we attempt more complex maneuvers, involving concurrent sideslip angle and angle of attack variation. As seen in Section 6.2, the yaw-axis QNPAS capability of the aircraft in an open-loop context is significantly less restricted in range and faster in response time than the pitch-axis capability. Pitch-axis capability is thus anticipated to be a determining factor in the performance of multiaxis QNPAS motions. Simulating a small scroll-shaped target path demonstrates the capacity of this system for complex QNPAS. Taking airspeed $U_c = 25$ m/s, target parameters $\hat{\beta} = 0.4$, $\hat{\alpha} = 0.2$, $\alpha_0 = 0.2$, and $\Gamma = 0.3$ rad with an inboard dihedral constraint, we compute the trim state control paths and perform flight simulations for several oscillatory period values.

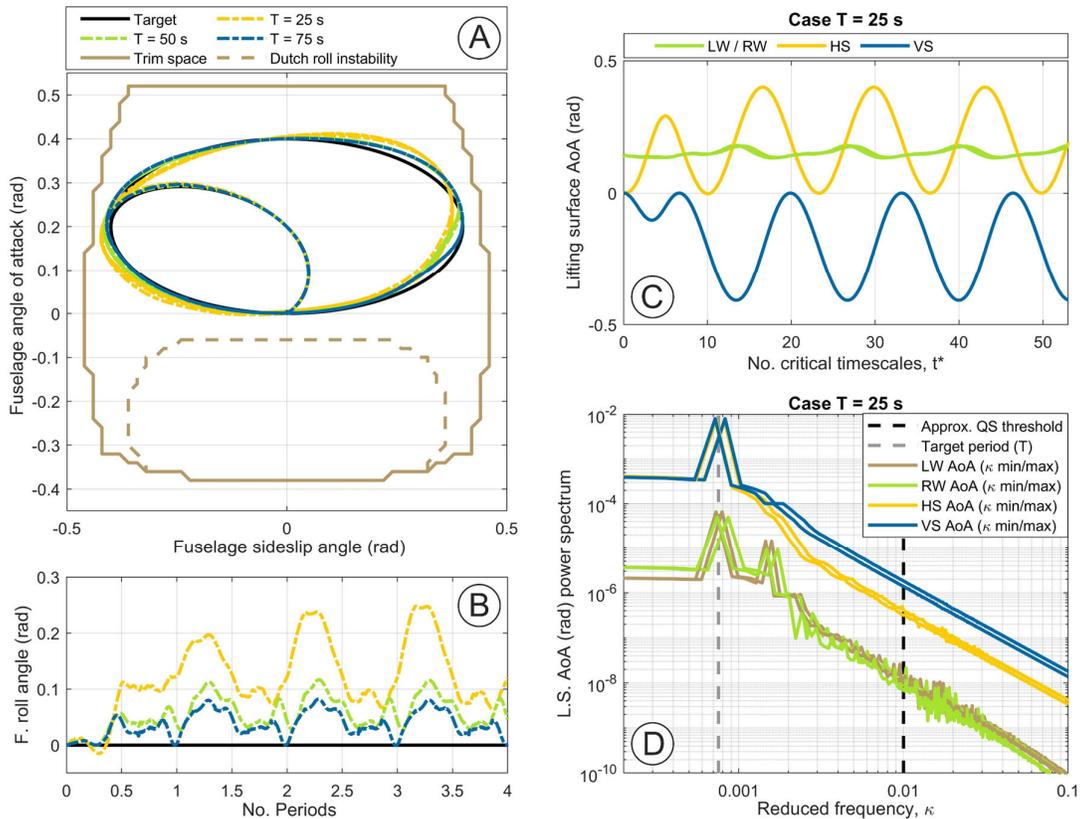

**Figure 11**: Simulation of successful scroll-shaped QNPAS. **A**: Target and response loci in aircraft (fuselage) angle of attack and sideslip. Note the symmetry in Dutch roll instability zone arising from switching outboard dihedral constraint. **B**: Aircraft roll response. **C**: Lifting surface tip angle of attack histories. **D**: Reduced frequency ($\kappa$) spectra of the lifting surface angles of attack, for minimum and maximum estimates of $\kappa$, and indicating the QNPAS target period and an approximate threshold for quasisteady aerodynamic model validity.

Page 23 of 32

Figure 11 shows the orientation results of these simulations, compared with the target paths. A good agreement with the orientation targets may be observed, and the magnitude of the observed discrepancies is only slightly affected by the period of the target path oscillation. Figure 12 shows a rendering of a QNPAS cycle for the $T = 25$ s maneuver, alongside the maneuver flight path and control histories (the latter, independent of $T$). The observed discrepancies may be attributable to the spiral dynamics of the aircraft (as seen in Figure 12) and/or the presence of gyroscopic torques engendered by multiaxis motion. In practical terms, this maneuver demonstrates a useful form of QNPAS in biomimetic UAVs – the ability to cycle through a biaxial path through the trim space. This capability can be utilized to orient fuselage-mounted equipment, and is a defined form of supermaneuverability, according to Herbst [6]: the first such form demonstrated in biomimetic aircraft. It is available at low thrust – no greater than 20% of gravitational acceleration ($0.2g$ w.r.t. the 8 kg aircraft).

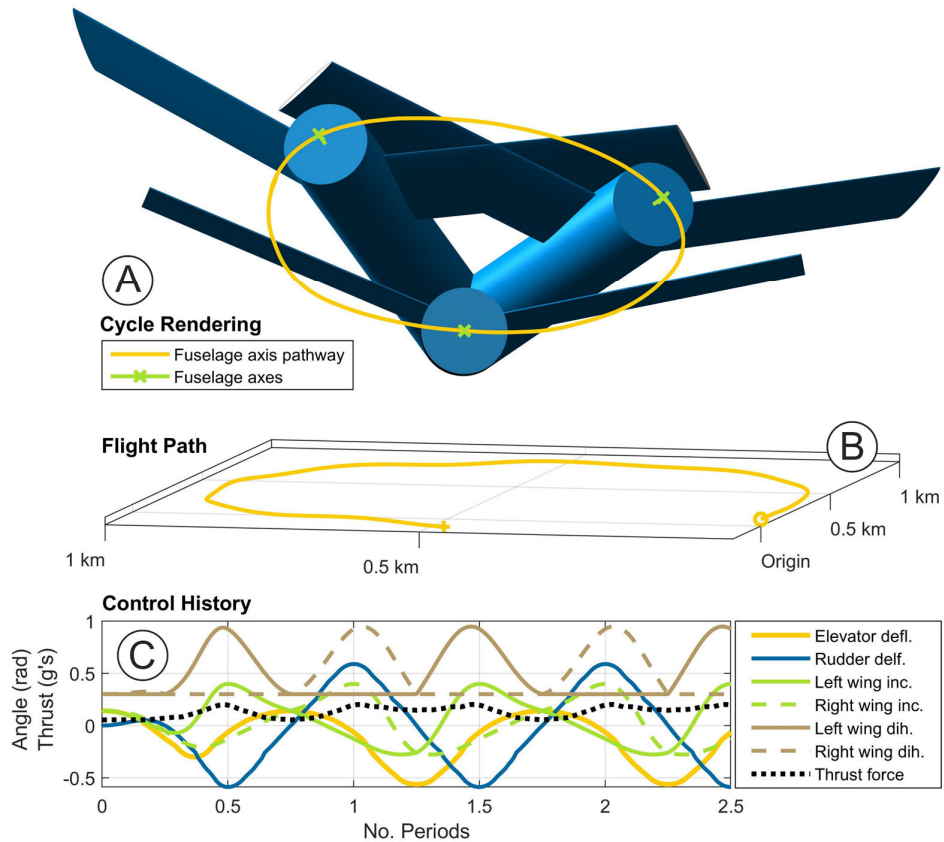

**Figure 12:** Simulation of successful scroll-shaped QNPAS. **A**: Rendering of the case study UAV QNPAS response locus: orientation w.r.t. flight velocity. **B**: Aircraft flight path. **C**: Control histories. Note the independence of control profiles to the target period ($T$): this is a property of the trim state guidance strategy.

### 7.2. Non-differentiable multiaxis QNPAS

The scroll-shaped trim state target path utilized in Figure 11 is continuously differentiable. The use of non-differentiable paths is anticipated to decrease the accuracy of the direct force control, due to the finite system response time. Figure 13 shows direct force control results for a rectangular trim state path, with leftwards ($l$), rightwards ($r$), upper ($u$) and lower ($b$) bounds and initialization path from $(\alpha_{tg}, \beta_{tg}) = (0,0)$ to $(0,u)$. As may be seen, the system performance for this nondifferentiable path is significantly worse than for the continuously differentiable scroll-shaped paths, with transient oscillations at discontinuities. Overall, however, the system response for the largest period ($T = 75$ s) is notably accurate.

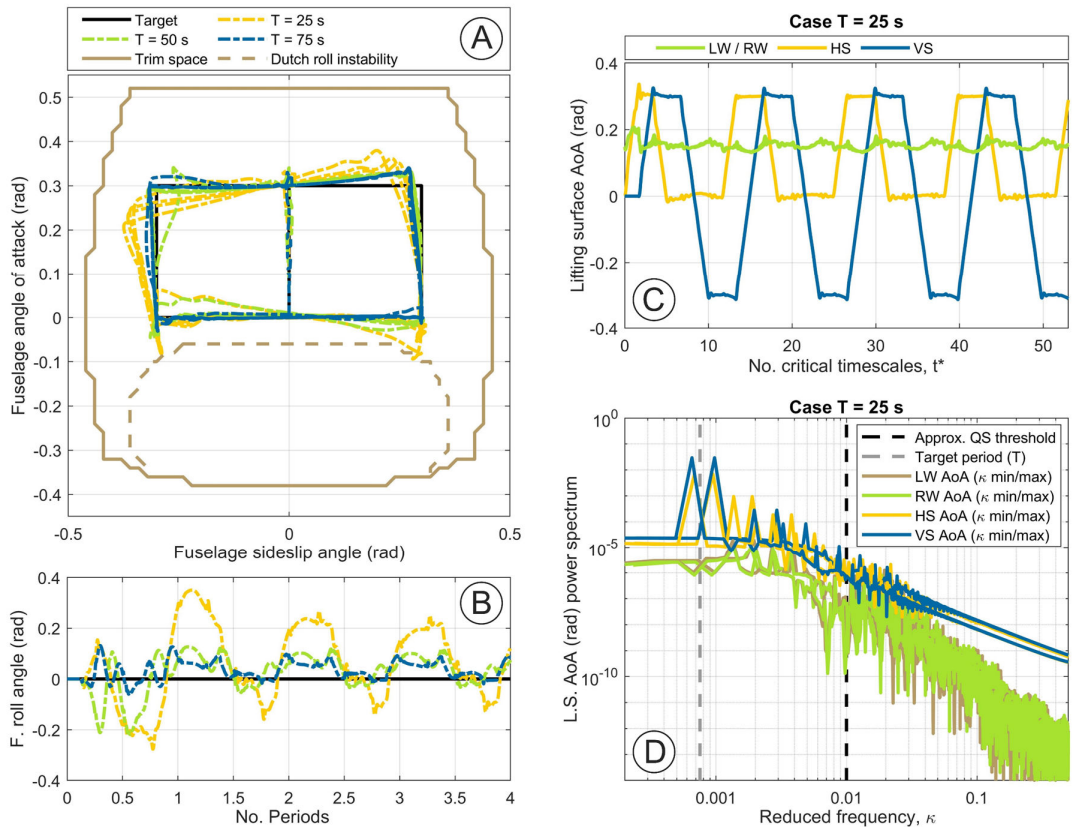

**Figure 13**: Simulation of successful nondifferentiable-path QNPAS. **A**: Target and response loci in aircraft (fuselage) angle of attack and sideslip. **B**: Aircraft roll response. **C**: Lifting surface tip angle of attack histories. **D**: Reduced frequency ($\kappa$) spectra of the lifting surface angles of attack, for minimum and maximum estimates of $\kappa$, with dashed lines indicating the QNPAS target period and an approximate threshold for quasisteady aerodynamic model validity.

### 7.3. QNPAS: Towards RaNPAS

The studies in Sections 6 and 7.1-7.2 are concerned with finding the minimum feasible period for a variety of QNPAS maneuvers – the logic being that larger periods allow greater time for the aircraft to stabilize at/to the local QNPAS target state. The strength of the stabilization effect is governed by the flight stability of aircraft; both static and dynamic. The estimates of the aircraft response time in Section 6 represent a broad proxy estimate of overall static and dynamic stability in independent pitch and yaw axes. The stability analyses of Section 5 provide charts of localities in the trim space where static or dynamic instability is expected. It follows that more rapid NPAS maneuvers through dynamically unstable regions of the trim space might be unsuccessful; both because of the dynamic instability, and because tracking-stabilization effects are not strong enough. The reverse is found to be the case. Not only are rapid NPAS maneuvers through unstable regions of the trim space successful; but in these regions, decreasing NPAS timescale – to levels far more rapid than the expected pitch response time of the aircraft – yields more accurate open-loop tracking response. Figures 14-15 shows these results, for two systems with outboard positive-dihedral ($\Gamma = 0.3$ rad) or inboard zero-dihedral ($\Gamma = 0$) constraints, under a scroll-shaped target path with $\beta_{\text{amp}} = 0.4$, $\alpha_{\text{amp}} = 0.3$, $\alpha_0 = 0$ rad, and $T \in [5,10,15]$ s.

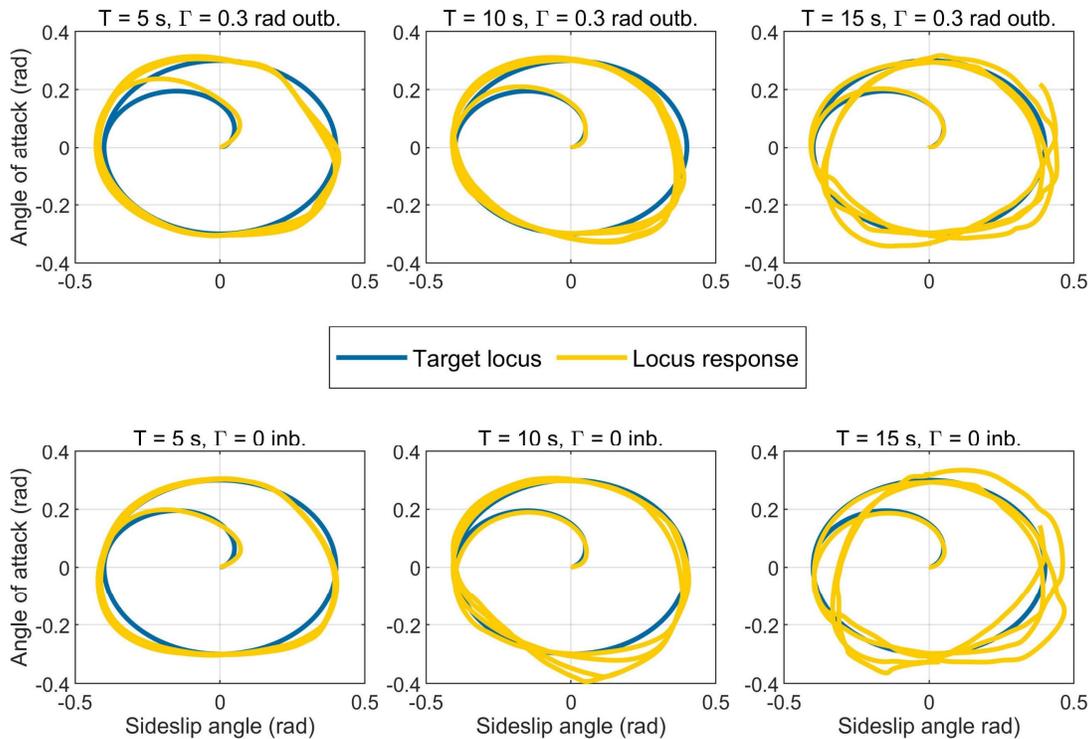

**Figure 14**: Target and response loci in aircraft (fuselage) angle of attack and sideslip for six different large-amplitude rapid QNPAS paths: periods $T = 5, 10, 15$ s; for configurations $\Gamma = 0.3$ rad outboard constraint (outb.), and $\Gamma = 0$ inboard constraint (inb.).

Page **26** of **32**

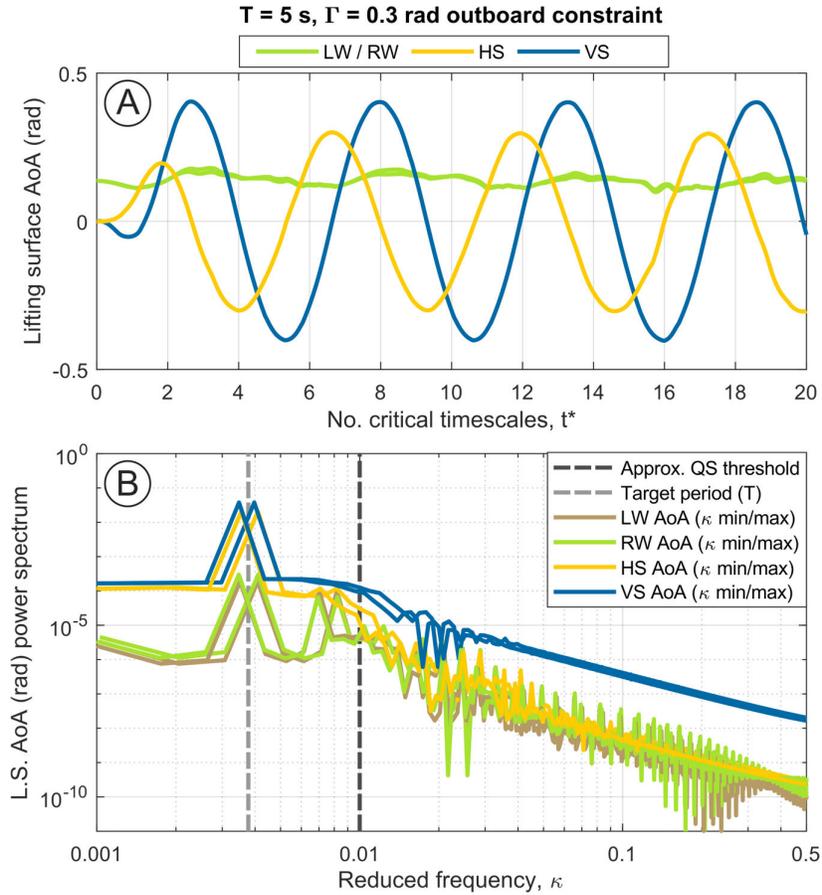

**Figure 15**: Lifting surface angle of attack results for a representative rapid QNPAS maneuver ($T = 5$ s, $\Gamma = 0.3$ rad outboard constraint). **A**: Time histories w.r.t. the critical quasisteady timescale $t^*$. **B**: Reduced frequency ($\kappa$) spectra, for minimum and maximum estimates of $\kappa$, with dashed lines indicating the QNPAS target period and an approximate threshold for quasisteady aerodynamic model validity.

There appear to be several interrelated effects at work. Firstly, independent of the presence of dynamic instability in the system, there may be an entrainment mechanism between pitch- and yaw-axis QNPAS motion – whereby the more rapid yaw tracking response entrains the slower pitch response in multiaxis motion. This could be responsible for general forms of response stability at timescales lower than the independent pitch response time. Secondly, with regard to the resilience of the system to Dutch roll instability, three possibilities are presented. (**1**) Switching rapidly between stable and unstable states may suppress the growth of instability. (**2**) The induced flow from rapid QNPAS motion may alter the stability of the Dutch roll mode. (**3**) For this particular QNPAS path, low-level Dutch roll instability may reinforce rotational motion: the Dutch roll period of is $T \cong 4$ s.

Further studies are required to address this fully – not only because of the interrelated effects at work; but because these more rapid simulations are beginning to reach the validity limits of our model. Figure 15 shows the lifting surface angle of attack time histories and spectral data for the simulation at $T = 5$ s. The spectral content above the unsteadiness threshold is of sufficient magnitude that unsteady aerodynamic effects may significantly alter the system behavior. These maneuvers would thus more properly be considered RaNPAS [4]: a capability which comes with its own, more complex, phenomena. The simulations of Dutch roll suppression are thus only representative of possible stabilization effects independent of more unsteady aerodynamic phenomena.

**7.4. Implications of QNPAS**

The studies in this section have demonstrated a range of QNPAS capability in the simulated case study UAV. This capability has several implications. It serves a general demonstration of this capability in analogous biomimetic aircraft: the mechanisms of QNPAS capability are simple (Section 4), and there is no reason to suspect that complex flow effects or changes in system parameters significantly alter its nature. The analysis of specifically low-transience NPAS behavior has dual implications: while the maneuvers studied are not yet as rapid as well-known forms of RaNPAS (e.g., Pugachev's cobra [2], and the hook maneuver [1]); they represent a form of NPAS behavior that can be maintained indefinitely, without reliance on unsteady aerodynamic or inertial effects. Further study into biomimetic RaNPAS is required, but prospects are hopeful, and initial steps have been taken: the capability studied in Section 7.3 is already a basic form of RaNPAS.

While the restriction of this analysis to a case study UAV means that details of the relationship between airframe configuration and QNPAS capability cannot be described in detail, a few aspects may be highlighted. The amplitude of QNPAS control available in these systems is limited by the control effectiveness of either the morphing wings, or the stabilizers. In the case study UAV, lack of stabilizer control effectiveness accounts for the entire amplitude limit on the trim space – and these stabilizers are not unrepresentatively small. This highlights a general need to ensure high stabilizer control effectiveness, in addition to morphing control effectiveness, in biomimetic systems with intended QNPAS capability. In terms of wing morphing control, the key degrees of freedom are the wing dihedral and incidence. For pitch-axis QNPAS control, symmetric incidence control will likely always be the most practical option. Additional yaw-axis control is best achieved via asymmetric incidence control, alongside asymmetric dihedral control. While differing forms of asymmetric dihedral control may confer various benefits, at simplest, only single-wing dihedral control is required. This level of wing rotation control (total 3-DOF) is entirely feasible for implementation in a light UAV. As such, it may have applications in artificially-intelligent UAV systems for dogfighting [17]: such systems may be able to significantly increase their dogfighting performance via these simple forms of biomimetic control.

## 8. Conclusion

In this work we have shown how biomimetic morphing-wing UAVs are capable of multiaxis nose-pointing-and-shooting (NPAS) behavior: the first form of classical supermaneuverability shown to exist in biomimetic aircraft. We devised and validated a nonlinear flight dynamic model for a case study biomimetic morphing-wing UAV. This model was then used to demonstrate the existence of a multidimensional space of stable trim states in biomimetic aircraft of this form. By navigating the aircraft morphing configuration through this space of trim states, multiaxis quasistatic NPAS maneuvers can be performed. These maneuvers are successful under open-loop control in simulation, and robust with respect to changes in roll and the presence of spiral mode instability. We explored the stability properties of the space of trim states, and the mechanisms and limitations of this form of NPAS motion. The interactions between NPAS and flight-dynamic stability are complex: the choice of morphing configuration allows a choice of stability properties, and some forms of motion interact with flight-dynamic instabilities – evidence of the suppression of Dutch roll instability is presented. Not only do these results demonstrate the range and variety of quasistatic NPAS capability available through biomimetic wing control; they also lay the groundwork for the study of rapid nose-pointing-and-shooting (RaNPAS) and more complex air combat-relevant supermaneuverability in these biomimetic systems.